\pdfoutput=1
\documentclass{article}
\usepackage{ijcai22}

\usepackage{times}
\usepackage{soul}
\usepackage{url}
\usepackage[hidelinks]{hyperref}
\usepackage[utf8]{inputenc}
\usepackage[small]{caption}
\usepackage{graphicx}
\usepackage{amsmath}
\usepackage{amsthm}
\usepackage{booktabs}
\usepackage{algorithm}
\usepackage{algorithmic}
\title{Efficient Symptom Inquiring and Diagnosis via Adaptive Alignment of Reinforcement Learning and Classification}
\author{
Hongyi Yuan$^1$ $^2$\and
Sheng Yu$^1$ $^2$\footnote{Contact Author}\and\\
\affiliations
$^1$Center for Statistical Science, Tsinghua University, China\\
$^2$Department of Industrial Engineering, Tsinghua University, China\\
\emails
yuanhy20@mails.tsinghua.edu.cn,
syu@tsinghua.edu.cn
}

\begin{document}

\maketitle

\begin{abstract}

Medical automatic diagnosis aims to imitate human doctors in real-world diagnostic processes and to achieve accurate diagnoses by interacting with the patients. The task is formulated as a sequential decision-making problem with a series of symptom inquiring steps and the final diagnosis. Recent research has studied incorporating reinforcement learning for symptom inquiring and classification techniques for disease diagnosis, respectively. However, studies on efficiently and effectively combining the two procedures are still lacking. To address this issue, we devise an adaptive mechanism to align reinforcement learning and classification methods using distribution entropy as the medium. Additionally, we created a new dataset for patient simulation to address the lacking of large-scale evaluation benchmarks. The dataset is extracted from the MedlinePlus knowledge base and contains significantly more diseases and more comprehensive symptoms and examination information than existing datasets. Experimental evaluation shows that our method outperforms three current state-of-the-art methods on different datasets by achieving higher medical diagnosis accuracy with fewer inquiring turns.
\end{abstract}

\section{Introduction}
Medical automatic diagnosis (MAD) is one of the main areas of research in the application of artificial intelligence in healthcare. An MAD system imitates a human doctor and interacts with the patients, collecting their symptom information and making confident diagnoses. Currently, people perform online searches for self-diagnosis. However, many results obtained from these searches may be inaccurate or irrelevant. Moreover, in real clinical encounters, diagnosis accuracy can be low in many rural areas due to limited healthcare resources. Therefore, a well-developed MAD system has great potential in supporting clinical diagnostic decisions, increasing diagnosis accuracies, and helping with the patients’ self-diagnosis. 
% MAD systems have great potential in simplifying the diagnostic procedure, increasing diagnosis accuracies and reducing patient costs.

Previous studies mainly utilized structured or unstructured medical data to develop a disease classification system. A large variety of machine learning or deep learning methods have been used to build such systems \cite{kononenko_machine_2001,mullenbach-etal-2018-explainable}. 
% For example, \cite{choi_doctor_2016} and \cite{choi_retain_2017} developed recurrent neural network (RNN)-based methods to perform differential diagnosis given the patients’ medical histories and current conditions. \cite{hosseini_heteromed_2018} used graph neural networks for disease diagnosis based on medical histories and knowledge graphs. 
% On TREC-CDS tasks, few recent works applied deep RL to retrieve medical disease concepts from clinical narratives with external knowledge bases (e.g., Wikipedia) \cite{ling_diagnostic_2017,ling_learning_2017}. 
However, these methods can only make diagnoses given rich medical histories. Doctors may meet patients with no clinical history and minimal initial symptom information in real clinical settings. These methods will fail in such scenarios.

In the real clinical setting, a medical diagnosis process can be formulated as a sequential decision-making problem \cite{yu2020reinforcement}. In the role of the doctor, the MAD system collects patients' symptom information by posing symptom-related queries or assigning medical examinations during a multi-round interaction with the patients. When the system collects enough information, it gives a final diagnosis of the disease. While more information is always beneficial for diagnosis accuracy, asking too many questions lowers the patients' experience, and taking too many examinations raises the medical costs. The goal of MAD is to achieve high diagnosis accuracy and reduce the costs of collecting symptom information. 

Some studies developed dialogue systems to interact with patients and give more personalized medical instructions \cite{li_semi-supervised_2021,lin_graph-evolving_2021}. These methods focused on natural language processing and could only provide a one-step dialogue interaction with the patients based on real dialogue histories. Our work does not aim to generate comprehensive dialogues or one-shot instructions but to approach the inner decision-making logic of real diagnostic processes to achieve highly effective and efficient automatic diagnosis.

With the promising capacity of reinforcement learning (RL) in solving sequential decision-making problems in various fields, some works developed symptom checkers for online healthcare services by RL. Tang \textit{et al.}~\shortcite{tang2016inquire} divided the diseases and related symptoms into 11 anatomical parts and applied deep Q-learning (DQN) \cite{DBLP:journals/corr/MnihKSGAWR13} methods on each part to inquire about symptoms. Wei \textit{et al.}~\shortcite{wei-etal-2018-task} also used DQN methods to inference diagnosis. Kao \textit{et al.}~\shortcite{Kao2018ContextAwareSC} and Liao \textit{et al.}~\shortcite{liao2020taskoriented} used hierarchical RL: a master agent is assigned to decide which anatomical part agent of the lower hierarchy to inquire about symptoms and when to give the diagnosis. Kao \textit{et al.}~\shortcite{Kao2018ContextAwareSC} also added contextual information of a patient (e.g., sex and age) to enhance the diagnosis accuracy through a Bayesian approach. Splitting diseases into several anatomical parts can reduce the task complexity. With appropriate joint policy training, the above works outperform naïve DQN methods. Xu \textit{et al.}~\shortcite{xu_end--end_2019} proposed a method called KR-DQN that embedded entity relations into a DQN agent. A trainable correlation matrix was added to the agent outputs to help inquire about related symptoms and give diagnoses. Peng \textit{et al.}~\shortcite{NEURIPS2018_b5a1d925} proposed a method based on the policy gradient\cite{REINFORCE} method, called REFUEL. They designed auxiliary rewards and used a dual neural network architecture rebuilding the sparse features of the patients to help discover the positive symptoms more quickly. 

All of the above works simply combined symptom inquiring and disease diagnosis into the same action space. This unified framework neglects the difference between these two kinds of actions. Symptom-inquiring actions aim to obtain informative symptoms with as few steps (or other kinds of costs) as possible, while the diagnosis action is one-step and aims to obtain an accurate diagnosis at the end of each diagnostic process. The former actions may fall into the RL framework, while the latter is naturally a classification problem. Forcing diagnosis as part of MDP increases the search space for the agent, limiting not only the diagnosis accuracy but also the symptom inquiring efficiency. In addition, they ignored the fact that negative symptoms also contain rich information and are critical in accurate clinical decision-making.

More recent works used a bipartite framework and solved the two kinds of actions separately. They used RL methods to inquire about symptoms and employed another disease classifier to give diagnoses given the current symptom information. A variety of methods were proposed to integrate RL agents and classifiers. Xia \textit{et al.}~\shortcite{xia_generative_2020} used a generative adversarial network \cite{goodfellow_generative_2014} framework to balance symptom inquiring and disease diagnosis, and also designed entropy-based rewards to guide the agent to inquire informative and rational questions. Their method trivially gives the diagnosis when the maximum inquiring step length is reached, or the entropy of the disease distribution is smaller than a fixed threshold. The threshold was chosen intuitively, and all diseases shared the same threshold. However, a proper threshold may differ across datasets or even diseases. Lin \textit{et al.}~\shortcite{lin2021towards} proposed a method called INS-DS comprising two cooperative modules: a symptom inquiry module and an introspective module. The introspective module intervenes in the potential responses of the inquiry and decides to give the diagnosis if the diagnoses of these interventions remain unchanged. \textit{Lin et al.}~\shortcite{lin_learning_2020} designed a criterion that the model gives the diagnosis when the probability of the preferred disease is beyond the upper bound of the $6\sigma$ confidence interval of the other diseases’ probabilities. The confidence interval is estimated by bootstrapping the classifiers, which would be computationally costly. Although they achieve good automatic diagnosis performance, the above methods are only tested on datasets containing few diseases or symptoms. These methods may fail or be inefficient when evaluated on complicated situations with various diseases and symptoms. 
% Moreover, the above three works only tested their method on datasets with few diseases, while there are hundreds of diseases and symptoms in real clinical settings. 

On this diagnostic decision-making task, a more recent work proposed a non-RL competitive bipartite framework, called FIT, using a multi-modal variational autoencoder model and a two-step sampling strategy for disease prediction \cite{he2021fit}. They reported state-of-the-art results on different datasets.

In our work, we handle two kinds of actions separately, performing the symptom inquiring part using RL and treating the diagnosis part as a classification task. To incorporate both parts, we monitor the entropy of disease distribution at each interaction turn and assume different entropy thresholds for each disease. Finding a proper threshold for each disease is critical to the performance of the MAD system, and it is impossible to set thresholds manually without prior knowledge. To address this problem, we propose an adaptive threshold approach which learns disease-exclusive thresholds along with training.

% we formulate adaptive thresholds for each disease and design an approach to learn the thresholds when training.

% We design a subtle approach to incorporate two models. In real diagnostic processes, a human doctor will arrive at a conclusion when the information collected provides enough confidence and rules out other possibilities. In information theory, entropy is a measurement of uncertainty. Therefore, in our work, we monitor the entropy given by a disease classifier. When the entropy is lower than a threshold, the symptom inquiring process will be terminated, and a diagnosis will be given. However, the entropy threshold is difficult to determine given different disease distributions, and a poor choice of the threshold will result in unsatisfactory performance of the MAD system
% % \footnote{We illustrate this through experiments shown in Appendix A1.}
% . To address this problem, we propose an adaptive approach for selecting a proper threshold.

The training of RL methods has to explore the diverse actions given various patients' known symptom statuses, and it is unavailable to train RL methods on real patients or medical histories in consideration of safety and privacy. Thus, the previous works evaluate their methods on synthetic patients simulated by medical knowledge bases or medical dialogue histories. However, the shortage of previous works is that the existing datasets either contain few diseases or have incomplete symptom information. Thus, the simulated patients result in unrealistic conditions. Pertaining to this problem, we explore the public online medical knowledge bases and propose a new dataset for simulating more realistic patients. The proposed dataset contains various diseases and more comprehensive symptom information.

The main contributions of this work are as follows:
\begin{itemize}
\item We publish a new dataset for synthetic patients simulation. Compared to other widely used datasets, the proposed dataset contains more comprehensive disease information, and we include medical examinations for the first time.
\item We propose an simple but effective method to automatically adjust entropy thresholds that incorporate an RL agent and a classifier. The thresholds change along with the training of the RL agent and the classifier.
\end{itemize}
Through extensive experiments, we show the efficiency and superior performance of the proposed method compared to other MAD methods. We also demonstrate the efficiency and effectiveness of our threshold selection method.

\section{Method}

\begin{algorithm}[tb]
\caption{Our proposed method}
\label{alg:algorithm}
\textbf{Input}: Disease knowledge base $\mathcal{D}$, Initial threshold $K_{\text{init}}$\\
% \textbf{Parameter}: The max time step $T$, The initial threshold $K_0$.\\
\textbf{Output}: Optimal policy $\pi_{\phi^*}$, classifier $f_{\psi^*}$, threshold $K^*$

\begin{algorithmic}[1] %[1] enables line numbers
\STATE Initialize $\pi_\phi$ and $f_\psi$ with random weights $\phi$ and $\psi$.
\STATE Initialize the threshold $K=K^{\text{init}}$.
\STATE Generate synthetic patients using $\mathcal{D}$.
\REPEAT
\STATE Sample a patient with disease $d_{\text{True}}$ and generate initial state vector $s_0$.
\STATE Calculate the initial entropy $H_\psi (s_0)$.
\WHILE{$t\le T$}
\STATE Inquire a symptom $a_t\sim \pi_\phi (s_{t-1})$.
\STATE Interact with the patient and obtain the next state $s_t$ and reward $r_t$.
\STATE Give a disease diagnosis attempt $d^t$ and calculate the disease distribution entropy $H_\psi (s_t)$.
% \STATE Calculate the entropy difference reward $r_H^t$.
% \STATE Calculate the overall reward $r_t=\mu r_p^t+\nu r_H^t$.
\STATE Store a training sample $\left(s_{t-1},a_t,r_t,d_{\text{True}}\right)$.
\IF {$H_\psi (s_t)<K_{d^t}$} 
\STATE End the diagnostic process.
\ENDIF
\ENDWHILE
\STATE The final diagnosis $d_{\text{diag}}=d^t$.
\STATE Update the policy and classifier network parameter using collected samples.
% \STATE Update the classifier parameter by cross-entropy loss.
\IF {$d_{\text{diag}}=d_{\text{gold}}$}
\STATE Update threshold $K_{d_{\text{diag}}}$ by Equation (1).
% \ELSE
\ENDIF
\UNTIL{End of epochs}
\end{algorithmic}
\end{algorithm}

Generally, a MAD system refers to a system that includes two parts: 1) the natural language part that parses and analyses the input language of a patient and outputs the inquiry question to the patient; 2) the dialogue management part that decides which symptoms to ask about and when to give the diagnosis according to the current information of the patient. Our method focuses on the latter because extracting symptoms from the patient’s responses and generating questions are handled by natural language processing or templates and are not part of clinical decision-making.

% Figure \ref{fig1} illustrates the architecture overview of our proposed model, and 
The pseudo-code shown in Algorithm \ref{alg:algorithm} illustrates the algorithmic overview of our proposed method. The model consists of a symptom-inquiring agent, a disease-diagnosing module, and a stopping criterion. The symptom-inquiring agent is a policy network inquiring about the patient’s symptoms, given the symptoms already known by the model. The disease-diagnosing module is a neural network classifier that outputs a disease classification given the currently known symptoms. The stopping criterion sets adjustable thresholds for each disease and decides when to stop inquiring about symptoms and to give a diagnosis. At each turn, the disease-diagnosing module receives the current patient’s symptom information vector and predicts the current disease distribution. Then we use this disease distribution to check and adjust the stopping criterion.

% Then, the entropy of the distribution is calculated to determine whether to give the diagnosis. When the entropy is below a dynamic threshold, which means we already have enough information for accurate diagnosis, then the symptom-inquiring agent stops asking for more symptoms and a diagnosis will be given by the classifier. If a correct diagnosis is given, the dynamic threshold is updated by the entropy of the final disease distribution.

% \begin{figure*}[t]
% \centering
% \includegraphics[width=0.7\textwidth]{figure1mod2.png} % Reduce the figure size so that it is slightly narrower than the column.
% \caption{Overview of our proposed method.}
% \label{fig1}
% \end{figure*}

\subsection{Notations}

We denote the number of all symptoms as $N$ and the number of all diseases as $M$. We design our symptom-inquiring agent based on an RL framework. The symptom-inquiring part is a finite-horizon MDP with a state space $\mathcal{S}$, an action space $\mathcal{A}$, and a policy $\pi_\phi$. Each state $s\in\mathcal{S}$ is a sparse vector of length $N$, and each entry of the vector indicates the status of the corresponding symptom, \textit{positive}, \textit{negative}, or \textit{unknown}, encoded by $1$, $-1$, and $0$, respectively. Each action $a\in\mathcal{A}$ is a symptom to inquire about. The size of the action space is $N$. The disease-diagnosing part is formulated as a classification task. The classifier $f_\psi$ receives the current patient’s state $s$ as input and outputs a disease distribution denoted as $p_\psi(d|s)$ and $p_\psi(d|s)=f_\psi(s)$, where $d$ denotes the diseases. Then, the entropy $H_\psi(s)$ of $p_\psi(d|s)$ is calculated to check the stopping criterion, formulate the reward for the symptom-inquiring agent and adjust the entropy thresholds of each disease.

\subsection{Disease-Diagnosing Model}

The disease-diagnosing model is an MLP classifier. Given the patient’s state $s_t$ at time step $t$, the state vector is fed to the classifier to generate the disease distribution. We derive the current diagnosis attempt $d^t$ and the entropy of the disease distribution $H_\psi(s_t)$ for the distribution as follows:
\begin{align*}
    &H_\psi(s_t) = -\sum_{i=1}^M p_\psi(d_i|s_t)\log\left( p_\psi(d_i|s_t) \right), \\
    &d^t = \operatorname{argmax}_{d\in\{d_i, 1\le i\le M\}}p_\psi(d|s_t).
\end{align*}
The entropy $H_\psi(s_t)$ and the diagnosis attempt $d^t$ are used to check the inquiring stopping criterion. If the $H_\psi(s_t)$ is smaller than the threshold of $d_t$, the MAD system terminates inquiring about patients' symptoms and returns $d^t$ as the final diagnosis.

The MLP classifier is trained together with the RL agent in a supervised fashion. At each inquiry turn, the current state vector is paired with the true disease label to form a training sample $(s_t,d_{\text{True}})$. Then, we train the classifier using these collected samples with cross-entropy loss. Thus, the classifier is accustomed to predicting the disease with partial symptom information.

\subsection{Symptom-Inquiring Agent}

% The symptom-inquiring agent $\pi_\phi$ uses an MLP policy network. At time step $t$, the agent generates a symptom-inquiring action $a_t$ given the current patient’s state $s_t$. The objective of RL is to maximize the expected accumulated rewards $\mathrm{E}_{\pi_\phi}[\sum^T_{t=0}\gamma^tr_t]$, where $\gamma$ is the discount factor. The agent is expected to inquire about the most valuable symptoms for obtaining an accurate diagnosis. We adopt the policy gradient method to optimize the network parameter $\phi$.

The symptom-inquiring agent $\pi_\phi$ generates actions concerning the patients' undiscovered symptoms given the patients' current states. The agent is expected to inquire about the most informative symptoms for obtaining an accurate diagnosis. 

The objective of RL is to maximize the expected accumulated rewards $\mathrm{E}_{\pi_\phi}[\sum^T_{t=0}\gamma^tr_t]$, where $\gamma$ is the discount factor. To achieve an efficient symptom-inquiring agent, we design a composite reward of two parts: $r_p$ and $r_H$. The details of the reward design will be discussed in Section~\ref{reward}.
% $r_p$ is designed intuitively to guide the agent to inquire about less symptoms and pose queries for unknown positive symptoms; $r_H$ incorporate entropy gains between two successive action steps and stimulate the agent for posing queries with more informative feed-backs. The details of the reward design can be found in Appendix \ref{reward}.

We also introduce an entropy regularization term $H\left(\pi_\phi(a|s)\right)=-\sum_{i=1}^N\pi_\phi(a_i|s)\log\left(\pi_\phi(a_i|s)\right)$ to the objective function to help the symptom-inquiring agent explore high-reward actions and escape from local optima at the beginning of the training stage. The parameter updating rule with respect to the objective is:
\begin{equation*}
     \phi' = \phi + \alpha\left[\frac{R(s,a)\nabla_\phi\pi_\phi(a|s)}{\pi_\phi(a|s)} + \beta \nabla_\phi H\left(\pi_\phi(a|s)\right)\right],
\end{equation*}
where $\alpha$ is the learning rate, $\beta$ is a weight parameter and $R(s,a)$ is the accumulated reward given state $s$ and action $a$.

\subsection{Reward Design}
\label{reward}

As in real clinical encounters, asking about too many negative symptoms is a waste of time, and taking too many medical examinations is a waste of medical resources and brings additional cost and even physical harm to the patient. Hence, we give a negative reward for each query. If the query discovers an unknown symptom, we assign an additional positive reward for the agent. The magnitude of positive rewards depends on the feedback of the patients. The goal of collecting symptom information is to give confident diagnoses. When our method gives a correct diagnosis, we provide the agent a positive reward, while when the method gives an incorrect diagnosis or fails to diagnose within the maximum step length $T$, a negative reward is given. This part of the reward, denoted as $r_p$, is designed to incentivize the agent to inquire about unknown and positive symptoms, pose fewer queries, and give correct diagnoses.

An informative symptom is not necessarily positive. Negative symptom feedback can also lower the uncertainty of diagnosis. Such negative symptoms can help rule out some candidate diseases. In information theory, entropy measures the level of uncertainty of a distribution. Thus, we track the entropy given by the disease classifier $\psi(s_t)$ at each time step $t$. The amount of information acquired by a symptom query is quantified by the entropy difference in the two successive inquiring steps. We introduce this to the reward design to guide the model to consider the symptom queries with more potential information feedback. The entropy difference reward is formulated as:
\begin{equation*}
     r_H = \max((H_\psi(s_t)-H_\psi(s_{t+1}))/H_\psi(s_0), 0).
\end{equation*}
where $s_t$ is the patient current state and $s_{t+1}$ denotes the next state. $H_\psi(s_0)$ is the disease distribution entropy based on the patient’s initial self-reports. It serves as a normalizing term to make the entropy difference reward consistent across different patients because the entropy difference can vary from very large to small given different patients’ self-reports. The maximum operator is to avoid negative information gain in the early stage of training when the classifier is imperfect because the entropy should decrease monotonically as more symptom statuses are collected.

Thus, the overall reward for training the symptom-inquiring RL agent is a weighted combination of $r_p$ and $r_H$:
\begin{equation*}
     r = \mu r_p +\nu r_H,
\end{equation*}
where $\mu$ and $\nu$ are the weights of the two kinds of rewards. We run grid search on the two parameters, and find that  Different choices of $\mu,\nu$ did not bring significant difference to the final accuracy, and they all outperformed the baselines. In the main experiments, we use $\mu=1$ and $\nu=2.5$.

\subsection{Stopping Criterion}
After designing a symptom-inquiring agent and a disease classifier, the remaining challenge is to set a proper stopping criterion to decide when to stop posing queries to give diagnoses. Intuitively, the method shifts to giving diagnoses when the entropy of the disease distribution is lower than a threshold.
The threshold indicates the attainable uncertainty degree for a disease. Doctors may deal with hundreds of potential diseases in real clinical encounters, and across different diseases, such thresholds may vary vastly. Thus thresholds are difficult to determine manually. To address this issue, our method contains an adaptive design to automatically find the proper thresholds.

We regard the thresholds for each disease as tunable parameters. In the training stage, we initialize the thresholds with some values. For each training episode, the disease classifier network will give a disease diagnosis when the distribution entropy is lower than the current threshold of the disease or the preset max step $T$ is reached. If the given diagnosis is correct, we use the final disease distribution entropy to adjust the threshold of the diagnosis. Otherwise, the threshold remains unchanged. The threshold is updated according to the following equation, for a disease $d\in\mathcal{D}$,
% At the beginning of training, we randomly initialize a reasonable threshold. During each training episode, our method gives a diagnosis if the entropy is lower than the threshold or the preset max step $T$ is reached. If a correct diagnosis is given, the disease distribution entropy is used to adjust the threshold according to the following equation:
\begin{equation}
     K_d = \lambda K_d + (1-\lambda)H_\psi(s_{\text{fin}}),
\end{equation}
where $K_d$ is the adjustable threshold for the disease $d$, $s_{\text{fin}}$ denotes the final patients’ state vector, and $\lambda$ is a Polyak parameter controlling the speed of updating the threshold. As the training progresses and the classifier becomes more accurate, the disease distribution entropy will reflect the diagnosis confidence adequately. With the above updating scheme, the automatically learned thresholds will set proper uncertainty levels for MAD to transit from symptom inquiring to diagnosing.

Within the maximum inquiring step limit, only if the disease distribution entropy is small than the threshold. The thresholds will keep decreasing when training. Therefore in practice, when a correct diagnosis is given, the threshold will be updated when the difference between final entropy and the threshold is large enough (i.e., $|K_d-H_\psi(s_{\text{fin}})|>\varepsilon$).

In the inference stage, we fix the learned thresholds. In each step, the classifier gives a diagnosis attempt by selecting the disease with the greatest probability and the disease distribution entropy. The MAD system decides whether to terminate the diagnostic process and give the final diagnosis by comparing the entropy and the threshold of the predicted disease.

\section{Experiments}

\subsection{Datasets}
Due to the highly sensitive nature of medical data and strict laws on their use, it is difficult to access real-world datasets on patients’ symptoms, examinations, and diagnoses. Therefore, we evaluate our method using two medical knowledge sources, SymCat\footnote{http://www.symcat.com
} and MedlinePlus\footnote{https://medlineplus.gov}, to generate synthetic patients. A simulated synthetic patient contains one disease and a set of relevant symptoms or medical examinations. We note that there are several early works (e.g., \cite{xia_generative_2020,xu_end--end_2019,lin2021towards}) that have evaluated their methods on two public medical dialogue datasets, namely the MuZhi Medical Dialogue Dataset \cite{wei-etal-2018-task} and the Dxy Medical Dialogue Dataset \cite{xu_end--end_2019}. However, these two datasets only contain a small number of diseases (Dxy contains 5 diseases and MuZhi contains 4 diseases) and are not appropriate benchmarks for realistic patients simulations because the real clinical encounters are more complicated. We discuss the details of the patient simulation procedures and training parameters in the Appendix \ref{train detail}.

% \subsection{SymCat:go to appendix}
% We follow \cite{Kao2018ContextAwareSC} and \cite{NEURIPS2018_b5a1d925} in using the SymCat dataset to simulate patients. SymCat contains 801 diseases and 474 symptoms. For each disease in SymCat, there are corresponding symptoms, context information (e.g., gender, age) and their occurrence probability. The patient simulation procedure first uniformly samples a disease. Then, the symptoms are generated by performing a Bernoulli trial on each corresponding symptom using their individual probability, and one of the sampled symptoms is assigned to be the patient’s self-report. We also generate each patient’s context information, including sex and age. Ages are encoded into several binary values, each representing a non-overlapping range of ages. The sex and age ranges are generated using the probabilities in SymCat. The encoded context information is concatenated to the state vectors. To compare our method with other baselines, we randomly sample $200$, $300$, and $400$ diseases to form 3 different disease sets. Furthermore, SymCat also provides disease categories. We extract the diseases belonging to the ‘Common Disease’ category to form another disease set. For each disease set, we sample $10^6$, $10^5$ and $10^5$ synthetic patients for training, developing and testing. Under the above simulation settings, each synthetic patient has approximately 3 symptoms on average.

\subsection{Proposed Dataset: MedlinePlus}

The symptom information in the SymCat dataset is incomplete. Some diseases in SymCat do not have enough symptom information. For example, ‘esophageal cancer’ only has superficial and ambiguous symptom descriptions such as ‘fatigue’ and ‘vomiting’, and the marginal probability of the symptoms is very low. This will result in an abnormal situation where a synthetic patient with ‘esophageal cancer’ only has 1 or 2 uninformative symptoms. Moreover, the dataset has no disease-related medical examination information. Thus the patients simulated using SymCat are not sufficiently realistic for testing clinical MAD systems.

We propose a new dataset extracted from MedlinePlus to simulate more realistic clinical encounters. 
% MedlinePlus is a public medical knowledge base that contains many disease-related articles. Each article contains subsections about symptoms and medical examinations. We used BIOS\footnote{https://bios.idea.edu.cn/}, a publicly available medical ontology, to identify symptom and medical examination entities from the articles and manually checked the extracted entities. The final proposed dataset contains 893 diseases, 1252 symptom-related concepts, and 860 medical examinations related concepts. The dataset include medical examination information for the first time. More details of the dataset and its construction can be found in Appendix \ref{mdlintro}.
%% MedlinePlus dataset can be viewed as a small knowledge graph containing the relations between diseases, symptoms and medical examinations. The resulting dataset has 618 diseases, 604 symptoms and 505 medical examinations. Each disease has approximately 12. related symptoms and 5. related medical examinations. 
% More details of the dataset and its construction can be found in Appendix.
% \section{Introduction of MedlinePlus}
% \label{mdlintro}
MedlinePlus is a publicly accessible website containing health information. There exist various articles on different aspects of healthcare information (e.g., Drugs \& Supplements, Genetics, Medical Tests). We construct the dataset from the disease articles on the MedlinePlus dataset. The construction procedure is as follows.

\begin{enumerate}
    \item We primarily focus on the disease article. For the ease of identifying the symptom and medical examination entities from the articles, we screen out the disease articles that do not contain \textit{Symptoms} and \textit{Tests} sections. This results in 1267 articles on different diseases.
    \item We use the vocabulary in BIOS\footnote{https://bios.idea.edu.cn/}, a machine-learned comprehensive biomedical knowledge graph, to extract the symptom and examination entities for each disease by maximum forward matching. All the extracted entities can be mapped to a concept (CID) in BIOS.
    \item To maintain the high quality of the dataset, we constrain the extracted symptom-related entities to the \textit{Sign, Symptom, or Finding} semantic type in BIOS and the extracted medical examination-related entities to the \textit{Laboratory Procedure} and \textit{Diagnostic Procedure} semantic types. Finally, we manually check every extracted entity and remove those ambiguous entities such as \textit{CN00003338 Diagnose} and \textit{CN00080784 Laboratory Tests}.
    \item To ensure the richness of information for each disease, we ditch the disease with no medical examination concept and the number of symptom concepts less than 3.
\end{enumerate}

With this procedure, the final proposed dataset contains 893 diseases, 1252 symptom-related concepts, and 860 medical examinations related concepts. Each disease connects to 13.61 concepts on average, where 8.31 are symptom-related, and 5.30 are medical examination-related. Our dataset includes disease-related medical examination information for the first time, providing more comprehensive symptom-related information. The dataset include medical examination information for the first time. 
% One drawback of our proposed dataset is the lack of disease prevalence and positive probabilities of a symptom given different diseases.

% \subsection{Training Details::GO TO APPENDIX}

\begin{table*}[h]
\centering
\resizebox{2\columnwidth}{!}{
\begin{tabular}{l|cc|cc|cc|cc|cc}
\hline
     & \multicolumn{2}{c|}{SymCat 200} & \multicolumn{2}{c|}{SymCat 300} & \multicolumn{2}{c|}{SymCat 400} & \multicolumn{2}{c|}{SymCat Common} & \multicolumn{2}{c}{MedlinePlus} \\
Model  &Accu.(\%) &Step$\downarrow$ &Accu.(\%) &Step$\downarrow$	&Accu.(\%) &Step$\downarrow$	&Accu.(\%) &Step&Accu.(\%) &Step$\downarrow$\\
\hline
Ours    &$\textbf{68.8}_{\pm 0.4}$  &$8.98_{\pm 0.01}$ &$\textbf{60.4}_{\pm 0.2}$ &$9.39_{\pm 0.01}$&$\textbf{55.5}_{\pm 0.1}$ 	&$10.02_{\pm 0.03}$&$\textbf{58.1}_{\pm 0.2}$	&$9.61_{\pm 0.01}$&$\textbf{96.2}_{\pm 0.1}$&$3.41_{\pm 0.02}$\\
REFUEL  &60.0&8.43&48.3	&9.65&33.8&10.18&47.8	&8.69&00.0&15.00\\
FIT\footnote{We use the results reported in their paper.}     &55.7&12.0&48.2	&13.1	&44.6&14.4&-&-&-&-\\
INS-DS  &46.1&\textbf{2.75}&39.6	&\textbf{4.46}&31.6	&\textbf{3.22}&36.2	&\textbf{2.98}&71.3	&\textbf{1.33}\\
\hline
\end{tabular}
}
\caption{Final diagnosis accuracies and inquiring turns on different datasets. The $\downarrow$ symbol indicates that the lower value is the better. For the results of our method, we run three times to calculate means and standard deviations.}
\label{table1}
\end{table*}

\begin{table*}[h]
\centering
\begin{tabular}{l|c|c|c|c|c}
\hline
& SymCat 200 & SymCat 300 & SymCat 400 & SymCat Common & MedlinePlus \\
        \hline
Ours    &$0.103_{\pm 0.001}$	&$0.108_{\pm 0.001}$	&$0.092_{\pm 0.002}$	&$0.100_{\pm 0.002}$	&$\textbf{0.234}_{\pm 0.001}$\\
REFUEL  &\textbf{0.167}	&\textbf{0.156}	&\textbf{0.145}	&\textbf{0.157}	&0.137\\
% FIT     &-	&-	&-	&-	&-\\
INS-DS  &0.011	&0.004	&0.008	&0.007	&0.002\\
\hline
\end{tabular}
\caption{Match rates on different datasets. For the results of our method, we run three times to calculate means and standard deviations.}
\label{table2}
\end{table*}

\subsection{Baselines}
We compare our method with three baselines, namely REFUEL \cite{NEURIPS2018_b5a1d925}, FIT \cite{he2021fit} and INS-DS \cite{lin2021towards}. REFUEL is a state-of-the-art RL method on SymCat. FIT achieves state-of-the-art accuracies on synthetic patients simulated by different datasets. INS-DS is a recently proposed method that splits the symptom-inquiring and disease-diagnosing actions and uses a different stopping criterion. INS-DS showed the best performance on the Dxy and MuZhi datasets. However, these three previous works did not provide open-source code. Thus, we re-implemented REFUEL and INS-DS according to the details described in their papers. FIT did not provide enough details for re-implementation. Therefore, we use the results reported in their paper for its performance.

\subsection{Main Results}
% \footnote{The case studies and the final threshold values are shown in Appendix A2 and A3.}}

% \begin{table*}[t]
% \centering
% % \resizebox{.95\columnwidth}{!}{
% \begin{tabular}{c|c|c|c|c|c}
%         & SymCat 200 & SymCat 300 & SymCat 400 & SymCat Common & MedlinePlus \\
%         \hline
% Ours    &\textbf{0.698}  &\textbf{0.615}  &\textbf{0.568}  &\textbf{0.588}  &\textbf{0.965}\\
% REFUEL  &0.600	&0.483	&0.338	&0.478	&0.000\\
% FIT    &0.557	&0.482	&0.446	&-	    &-\\
% INS-DS  &0.461	&0.396	&0.316	&0.362	&0.584\\
% \end{tabular}
% % }
% \caption{Diagnosis accuracies on different datasets.}
% \label{table1}
% \end{table*}

A good MAD system should achieve high diagnosis accuracy and pose as few symptom inquiries as possible. Thus, we use diagnosis accuracy and inquiring turns to evaluate our method. We show the results in Table \ref{table1}. 

On different disease sets from SymCat, our method achieves the best diagnosis accuracies, significantly outperforming other methods. 
Our methods use similar or fewer inquiring steps to achieve this high accuracy. Although the randomly selected 200, 300, and 400 disease sets are not exactly the same as those used in the FIT paper, our method surpasses FIT’s reported results by a large margin. 
INS-DS achieves the minimum inquiring turns, and its diagnostic accuracy is significantly lower than other methods. INS-DS tends to be over-confident about diagnosis and may give erroneous diagnostic decisions without collecting enough symptom information. 

% but the method actually inquires neither positive nor informative symptoms, and the patients’ self-reports mainly determine its accuracies.
% The failure of INS-DS is most likely due to the inappropriate reward design that places too much emphasis on the importance of diagnosis actions and gives no penalty for repeated inquiries. This reward design may be only suitable for small disease sets. 
% Another drawback of INS-DS lying in its stopping criterion is that the agent may inquire about an irrelevant symptom, and given the positive or negative state of the symptom, the classifier results in the same diagnosis. This circumstance may mislead the method to make a false diagnostic action; thus, IND-DS fails to complete the diagnostic process.

On MedlinePlus dataset, our method also outperforms all baselines in diagnosis accuracy. REFUEL fails to give diagnosis actions and keeps inquiring about symptoms until the maximum step. INS-DS gives diagnosis only based on patients' self-reports without collecting inquiring about symptoms carefully.

We also report the match rates used by previous works as another evaluation metric. The match rate is the average recalled positive symptoms per turn and measures the ability of the method to hit a positive symptom. The match rate results are listed in Table \ref{table2}. On MedlinePlus dataset, our method shows a great capability in discovering patients' symptoms, outperforming others. On different SymCat disease sets, REFUEL collects most patients' positive symptoms. The method still faces obstacles in increasing diagnosis accuracy with most positive symptom information. This reveals that it is difficult for a unified MAD framework to learn the difference and connection between symptom-inquiring and disease-diagnosing. An agent can quickly gain positive rewards from symptom-inquiring actions because a patient has many positive symptoms while possessing only one disease. INS-DS fails to discover any patients' positive symptoms on all datasets. The method fails when the decision space becomes larger. 

% INS-DS fails to hit any symptom. REFUEL achieves the best match rate results on SymCat datasets, showing excellent capability in inquiring about positive symptoms due to its feature rebuilding trick. REFUEL collects more patients' positive symptom information, but the method still faces obstacles in increasing diagnosis accuracy. As previously mentioned, combining symptom-inquiring and disease-diagnosing actions will enhance the difficulty in learning good behaviors of both action spaces. The agent may not learn the connection between two actions solely by maximizing pre-designed rewards, and the inquired positive symptoms may contain limited information for diagnosis. Because in MedlinePlus's large combined action space, a patient has many possible symptoms while possessing only one disease, the agent can more easily gain positive rewards from inquiries than diagnosis. The penalty against reaching the maximum step is relatively small. Through this inappropriate reward design, REFUEL does not learn to give diagnosis based on symptom information. It takes a conservative policy to obtain rewards by only acquiring symptoms until the step limits.

The experimental results show that the symptom match rate is a biased metric because it is not the purpose of designing a MAD system to discover patients' symptoms but to give accurate diagnosis. Besides, as explained previously, negative symptoms may also be helpful by ruling out other potential diseases. We verify this through a case study in a following case study.

\subsection{Case Study}

In this case, we present a diagnostic process accomplished by our method. The details of the case are in Table \ref{table:case}. The patient's self-reports, posed queries, symptom status feedback, and corresponding entropy changes are shown. We also show the proportions of each query existing among diseases. Through the scope of entropy, we can see that our model primarily poses queries of high existing proportions to narrow down the potential diseases and then poses queries shared among fewer diseases to reinforce the diagnosis judgment. Our model terminates the diagnostic process when the uncertainty for diagnosing is low (disease distribution entropy is only 0.043) and a correct diagnosis is given. Through this example, it is verified that negative feedback for a disease can also bring much information (i.e., a significant drop in entropy)
. After posing a \textit{CN00066522 Blood Tests} quires and getting negative feedback, the disease distribution entropy drops from 1.344 to 0.414. 

The table shows that our method tends to assign patients to various medical examinations that are regarded as advanced symptoms in our experiment settings. This phenomenon is consistent with the actual diagnostic processes. The medical examinations are more critical for patients' diagnosis than those superficial symptoms.

\begin{table*}[t]
\centering
% \resizebox{.95\columnwidth}{!}{
\begin{tabular}{c|cccc}
\hline
\textbf{Self-report}&\multicolumn{4}{c}{CN00022194 Cough}  \\
&\multicolumn{4}{c}{CN00090165 Joint Pain} \\
\hline
\textbf{Inquiring Turn}  & \textbf{Symptom} & \textbf{Proportion} & \textbf{Feedback}&\textbf{Entropy} \\
\hline
1  & CN00000560 Physical Exam& 358/893 &Positive& 2.075$\rightarrow$1.344\\
2  &CN00066522 Blood Tests& 256/893  &Negative& 1.344$\rightarrow$0.414\\
3  &CN00455502 Complete Blood Count &179/893  &Positive &0.414$\rightarrow$0.133 \\
4  &CN00015431 CT Scan& 181/893  &Negative &0.133$\rightarrow$0.099\\
5  &CN00002615 Chest X-ray &138/893 &Positive &0.099$\rightarrow$0.077\\
6  &CN00353823 Electrocardiogram &97/893 &Negative&0.077$\rightarrow$0.043 \\ 
\hline
\textbf{Diagnosis}&\multicolumn{4}{c}{Sjögren Syndrome}\\
\hline
\end{tabular}
% }
\caption{The diagnostic processes of disease \textit{Sjögren Syndrome}.}
\label{table:case}
\end{table*}

\subsection{Discussion}
% In this section, we show extensive experiment results to give thorough and detailed understanding of our proposed method.

\subsection*{Final threshold values}
The distributions of the final threshold values on different datasets are depicted in Figure \ref{fig2}. We use a box plot to illustrate the distribution of the threshold values of different diseases. Through the figure, we show that the threshold values' variance is large, relative to the value mean, indicating the significant threshold value difference between different diseases. Some diseases are potentially more distinguishable than others of which the final threshold values are lower. The final thresholds also can be used to identify the difficult diseases. The results illustrate that assigning different thresholds for different diseases is necessary to achieve differential diagnosis.
\begin{figure}[t]%{0.46\textwidth}
    \centering
    \includegraphics[width=0.9\columnwidth]{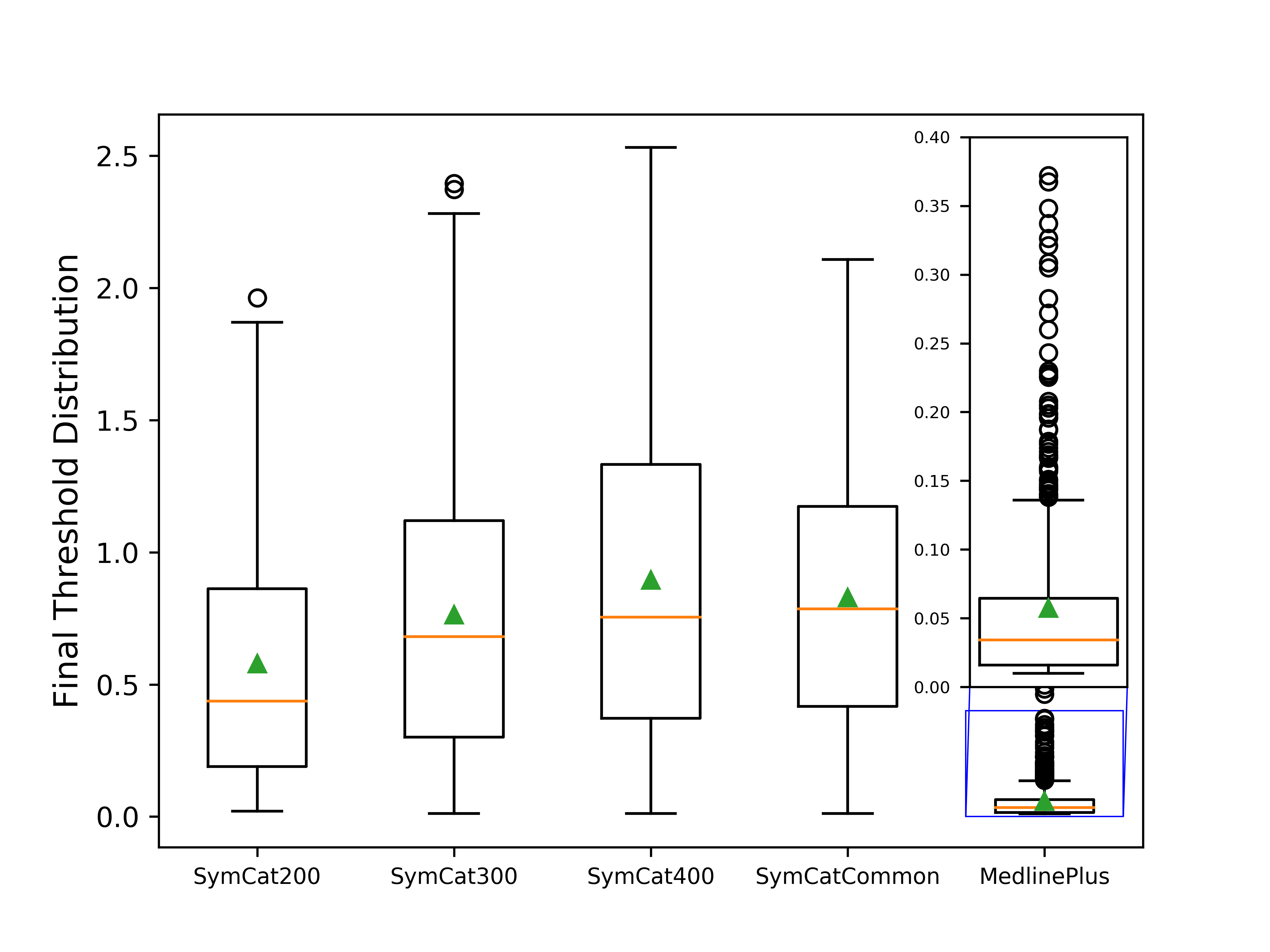}
    \caption{The distribution of the final thresholds on different datasets. The green triangles indicates the mean of the thresholds and the orange line indicates the median.} 
    \label{fig1}
\end{figure}%
%   \vspace*{\fill}

\subsection*{Comparing with unified and fixed thresholds}
\begin{table}[t]
\centering
\begin{tabular}{l|cc|cc}
\hline
 & \multicolumn{2}{c|}{SymCat 200} & \multicolumn{2}{c}{MedlinePlus} \\
Threshold &Step$\downarrow$ &Accu.(\%) &Step$\downarrow$ &Accu.(\%)  \\
\hline
Ours &8.98 &68.8 &3.41 &96.2 \\
Fix 2 &1.25&30.9&0.67&77.2\\
Fix 1 &6.23&64.1&1.49&86.9 \\
Fix 0.1 &10.77&71.7&3.31& 95.7\\
Fix 0.01 &11.55&69.7&7.34&88.5 \\
\hline
\end{tabular}
\caption{Comparing the proposed adaptive and disease-exclusive threshold approach with different fixed thresholds.}
\label{table3}
\end{table}

We investigate the effectiveness of the proposed stopping criterion with the adaptive threshold design. We conduct studies on the synthetic patients simulated on two datasets: our proposed MedlinePlus dataset and Symcat 200 dataset. Under the same experiment settings as the ones in the main results, we use different fixed thresholds in the stopping criterion, and across different diseases, the fixed thresholds are the same across diseases. The accuracy and inquiring step results are shown in Table \ref{table3}. On the larger scale dataset MedlinePlus, our adaptive thresholds achieve the highest diagnosis accuracy with a few inquiring steps. On the smaller scale dataset SymCat 200 disease, although the final diagnosis accuracy is not the highest using adaptive thresholds, the accuracy is comparable with the highest using less inquiring steps.

The results show the effectiveness of our adaptive threshold method. Using our method, we can find a good symptom inquiring step and diagnosis accuracy trade-off. The results of different fixed thresholds also validate that a suitable fixed threshold is hard to choose, and different choices of values profoundly impact the performance of the MAD system. This shows the practical importance of finding the thresholds automatically.

\subsection*{Robustness of Adaptive Threshold}
  
\begin{table}[t]
\centering
\begin{tabular}{c|cc|cc}
\hline
Init. Value &Mean &Std. & Accu.(\%) & Step$\downarrow$ \\
\hline
0.1 &0.059 &0.103 &96.1 &3.37 \\
1  &0.057 &0.101&96.2 & 3.41\\
2 &0.057 &0.105&96.1 &3.31\\
4 &0.059 &0.104&96.0 &3.37\\
Rand.1 &0.057 &0.097&96.0 &3.36\\
Rand.2 &0.059 &0.105&96.2 &3.41\\
\hline
\end{tabular}
\caption{The final adaptive thresholds of different initial values on MedlinePlus.}
\label{table4}
\end{table}
% \begin{figure}[t]%{0.46\textwidth} 
%     \centering
%     \includegraphics[width=0.9\columnwidth]{accu_ver2.png}
%     \caption{The diagnosis accuracy changing curves with different threshold starting values.}
%     \label{fig3}
% \end{figure}%

To further illustrate the robustness of our adaptive entropy threshold updating scheme, we conduct more experiments on the MedlinePlus dataset. A robust entropy threshold updating scheme is expected to maintain similar final thresholds values and diagnosis accuracies with different threshold initial values. In this experiment, we select different starting values of the entropy thresholds and repeat the experiment on the MedlinePlus dataset, keeping other training settings the same. We select 4 different starting values 0.1, 1, 2 and 4 which are unified across different diseases, and we also randomly assign different values between 0 and 1 to each disease as initial values. In Table \ref{table4}, we list the means and standard deviations of final thresholds and the corresponding accuracies. As shown in the table, with different initial values, the entropy thresholds and the accuracies reach similar final values. This shows that our threshold updating scheme is not sensitive to the selection of starting values and robustly finds the proper entropy thresholds.

\section{Conclusion}

In this work, we emphasize the critical importance of treating symptom-inquiring and diagnosis actions separately and incorporating both efficiently. We propose a new MAD method containing an RL symptom inquiring agent and a disease classifier and devise a simple adaptive threshold learning approach to align the two parts effectively. Our method sets new state-of-the-art results across different datasets, including our new proposed MedlinePlus dataset by achieving more accurate and more efficient automatic medical diagnosis. Our proposed dataset mitigates the problem that the current benchmarks are either too small-scale or consist of incomplete diagnostic information. Our proposed dataset contains more comprehensive and complete information on various diseases. Experimental evaluations on different datasets confirm the superiority and validity of our MAD system, and extensive experiment shows the necessity and robustness of the adaptive threshold learning approach.

% In this work, we propose a new MAD method that achieves higher accuracies with more efficient symptom inquiring. We set new state-of-the-art results across different dataset including our new proposed one. 
% Our method treats the symptom-inquiring and disease-diagnosing actions separately and devises a effective mechanism to cooperate two actions adaptively through the entropies of disease distributions. 
% % Our reward design helps the model learn to inquire informative and discriminative symptoms to get high diagnosis accuracies. 
% We also proposed a new dataset for simulating realistic patients. Our proposed dataset can serve as a better evaluating dataset, because it contains more comprehensive and complete information of diseases. Experimental evaluations on different datasets with multiple baseline methods confirm the superiority and validity of our proposed method that achieves accurate and efficient automatic medical diagnosis, requiring less training time.

\bibliographystyle{named}

\bibliography{draft.bib}

\newpage
\appendix
\section*{Appendix}

\section{Training Details}
\label{train detail}

\subsection{Network Architecture}

The policy network of symptom-inquiring agent is a 4 layers multi-layer perceptron (MLP) with hidden size (5120, 10240, 5120) on MedlinePlus and (2048, 2048, 2048) on SymCat disease subsets. The disease classifier for diagnosis is a 3 layers MLP with hidden size (5120, 5120) on MedlinePlus and (2048, 2048) on SymCat. All the activation functions used are GELUs \cite{hendrycks2020gaussian}.

\subsection{Parameters}

We train our model with the following settings. The maximum step length $T$ is set to 15. The cost of each inquiry is set to -1, an additional 0.7 reward is provided for unknown negative symptoms, and an additional 1.7 reward is provided for positive ones. These detailed reward numbers for negative and positive symptom feedback are designed intuitively, and slightly changing the proportion will have a marginal impact on the final result. The additional reward for a correct diagnosis is 1, while an incorrect diagnosis or failure to give a diagnosis until the maximum step limitation is given -1 as the reward. The coefficients of the reward weights $\mu$ and $\nu$ are set to 1 and 2.5, respectively. The weight parameters are decided by grid search. If not specified otherwise, the initial entropy thresholds for diseases are set to 1, the Polyak coefficient $\lambda$ of the updating threshold is 0.99 and $\varepsilon$ is set to 0.01. We tried different $\lambda$ such as 0.9, 0.99 and 0.999, and they resulted similar accuracies. We selected a compromised one 0.99. Different $\mu,\nu,\lambda$'s did not bring significant difference to the final accuracy, and they all outperformed the baselines. 

We train our symptom-inquiring agent in an on-policy manner. The adaptive thresholds and the policy and classifier network parameters are updated every 200 episodes. The collected samples are used for training both neural networks. The entropy threshold is updated using the average final entropies of the episodes given the correct diagnosis. We use Adam as the optimizer \cite{kingma2017adam}. The learning rate for the policy network is $2\times 10^{-5}$, and the learning rate for the disease classifier is $10^{-4}$. The discount-rate parameter $\gamma$ is set to 0.99, and the parameter $\beta$ for the entropy regularization term is set to 0.01 initially and gradually decreases to 0 overtime.

\subsection{Patients Simulation Procedure on Symcat}
\label{simulation}

For fair comparison, we follow \cite{Kao2018ContextAwareSC} and \cite{NEURIPS2018_b5a1d925} using the SymCat dataset to simulate patients. SymCat contains 801 diseases and 474 symptoms. For each disease in SymCat, there are corresponding symptoms, context information (e.g., gender, age), and their occurrence probability. The patient simulation procedure first uniformly samples a disease. Then, the symptoms are generated by performing a Bernoulli trial on each corresponding symptom using their individual probability, and one of the sampled symptoms is assigned to be the patient’s self-report. We also generate each patient’s context information, including sex and age. Ages are encoded into several binary values, each representing a non-overlapping range of ages. The sex and age ranges are generated using the probabilities in SymCat. The encoded context information is concatenated to the state vectors. To compare our method with other baselines, we randomly sample $200$, $300$, and $400$ diseases to form 3 different disease sets. Furthermore, SymCat also provides disease categories. We extract the diseases belonging to the ‘Common Disease’ category to form another disease set. We sample $10^6$, $10^5$, and $10^5$ synthetic patients for training, developing, and testing for each disease set. Under the above simulation settings, each synthetic patient has approximately 3 symptoms on average (3.0 symptoms on SymCat Common disease set, 3.4 symptoms on SymCat 400 disease set, 3.5 symptoms on SymCat 300 disease set, , 3.4 symptoms on SymCat 200 disease set).

\subsection{Patients Simulation Procedure on MedlinePlus}
\label{simulationmdl}

To simulate patients, we uniformly sample diseases from the disease set. Then, we sample symptoms for each disease, with the number of symptoms determined according to a Poisson distribution. Some of the sampled symptoms are assigned to patients’ self-reports. Medical examinations are treated as advanced symptoms of the diseases and are added to the patients’ overall symptoms. With this simulation process, each patient has on average 5.3 medical examinations and 6.5 symptoms, and 2.9 of the symptoms were self-reports. We simulate $10^5$ patients for each training epoch, and we simulate separately $10^5$ patients for development and testing.

\section{More details of MedlinePlus}
\label{mdlintro}

\subsection{Dataset Example}

\begin{figure}[t]%{0.46\textwidth} 
    \centering
    \includegraphics[width=0.9\columnwidth]{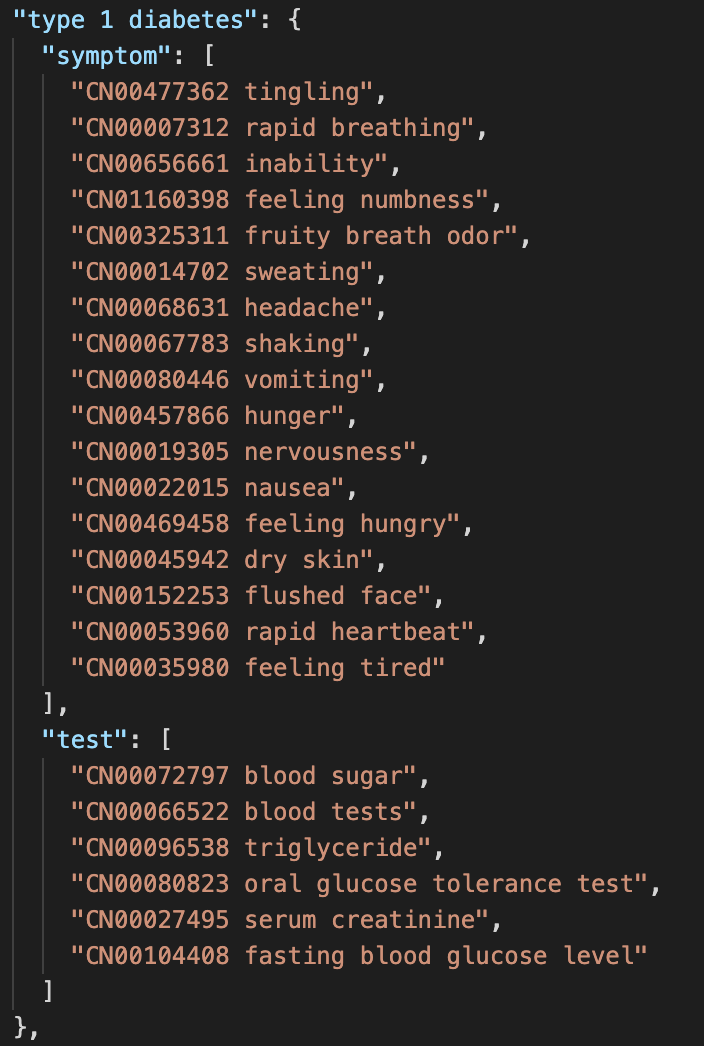}
    \caption{An \textit{Type 1 Diabetes} example in the proposed MedlinePlus dataset.}
    \label{fig2}
\end{figure}

The proposed MedlinePlus can be viewed as a medical knowledge graph for synthetic patient simulation. In the dataset, there are diseases in the string forms. For each of the diseases, there are related symptoms and medical examinations presented by \textit{CUI} connected to the disease. The whole dataset is provided in a json file in the supplementary materials along with the codes. We present an example disease \textit{Type 1 Diabetes} in Figure \ref{fig2}. 

\section{More Figures}

In this part, we show more directly visualized final adaptive threshold values on each dataset as supplements to the box plot in Discussion Section. In each of the following figures, the x-axis indicates the encoded disease indices in each dataset, and the y-axis shows the threshold value. Therefore, each blue spot represents the final adaptive threshold value for each disease. The red line indicates the average threshold value in each figure.

\begin{figure}[t]%{0.46\textwidth} 
    \centering
    \includegraphics[width=0.9\columnwidth]{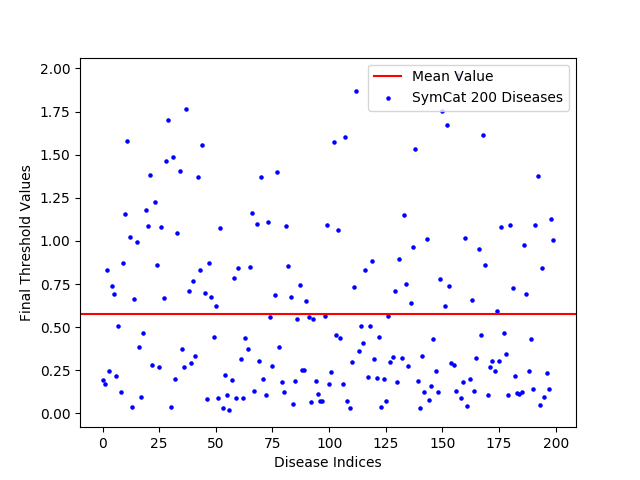}
    \caption{The final threshold value scatter plot of \textit{SymCat 200 Diseases} dataset.}
    \label{fig3}
\end{figure}%
\begin{figure}[t]%{0.46\textwidth} 
    \centering
    \includegraphics[width=0.9\columnwidth]{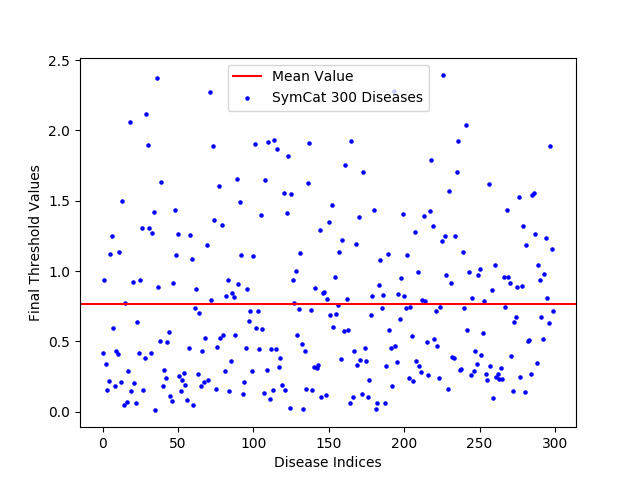}
    \caption{The final threshold value scatter plot of \textit{SymCat 300 Diseases} dataset.}
    \label{fig4}
\end{figure}%
\begin{figure}[t]%{0.46\textwidth} 
    \centering
    \includegraphics[width=0.9\columnwidth]{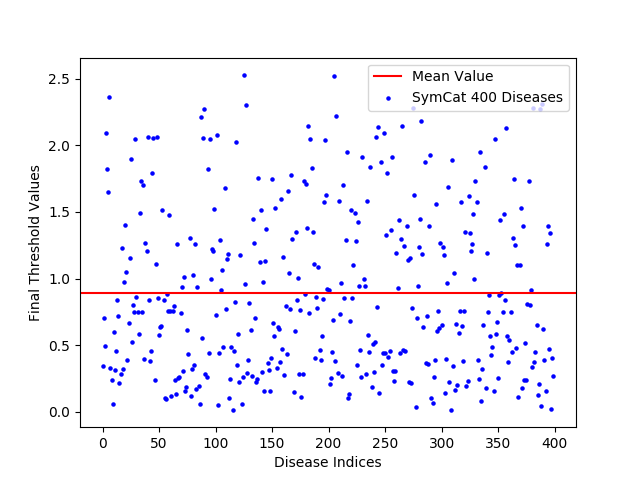}
    \caption{The final threshold value scatter plot of \textit{SymCat 400 Diseases} dataset.}
    \label{fig5}
\end{figure}%
\begin{figure}[t]%{0.46\textwidth} 
    \centering
    \includegraphics[width=0.9\columnwidth]{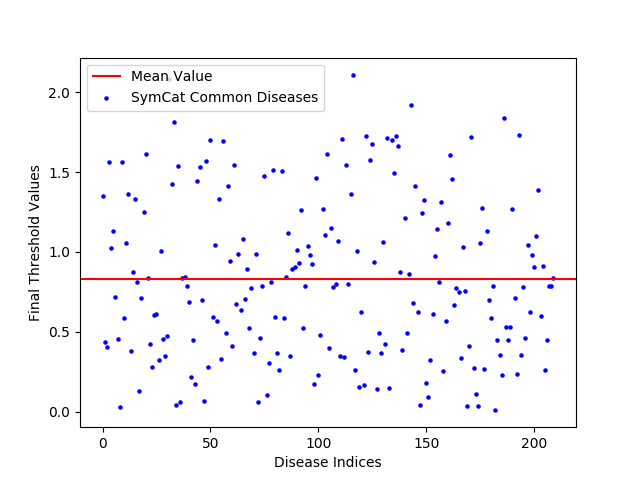}
    \caption{The final threshold value scatter plot of \textit{SymCat Common Diseases} dataset.}
    \label{fig6}
\end{figure}%
\begin{figure}[t]%{0.46\textwidth} 
    \centering
    \includegraphics[width=0.9\columnwidth]{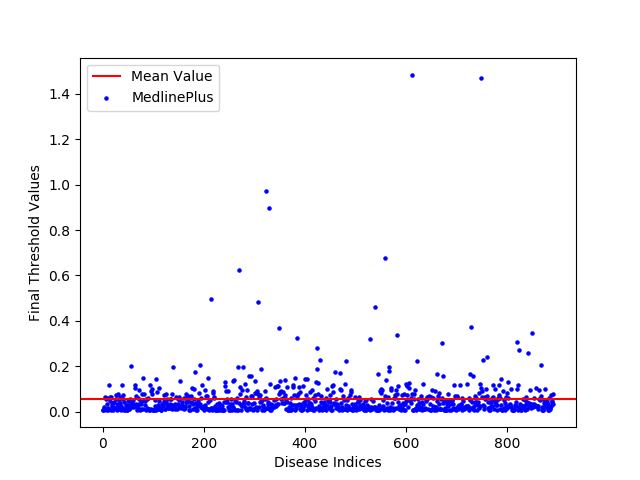}
    \caption{The final threshold value scatter plot of the proposed \textit{MedlinePlus} dataset.}
    \label{fig7}
\end{figure}%

\newpage
\newpage

\end{document}